\documentclass[10pt, conference, a4paper]{IEEEtran}
\IEEEoverridecommandlockouts

\usepackage{times}
\usepackage{epsfig}
\usepackage{graphicx}
\usepackage{amsmath}
\usepackage{amssymb}
\usepackage[super]{nth}
\usepackage[warn]{textcomp}
\usepackage{mathtools}
\usepackage[T1]{fontenc}
\usepackage[pagebackref=true,breaklinks=true,letterpaper=true,colorlinks,bookmarks=false]{hyperref}

\def\BibTeX{{\rm B\kern-.05em{\sc i\kern-.025em b}\kern-.08em
    T\kern-.1667em\lower.7ex\hbox{E}\kern-.125emX}}
\begin{document}

%%%%%%%%% TITLE
\title{Encoder-Decoder Based Convolutional Neural Networks with Multi-Scale-Aware Modules for Crowd Counting}

\author{\IEEEauthorblockN{Pongpisit Thanasutives}
\IEEEauthorblockA{Graduate School of Information\\ 
Science and Technology\\
Osaka University\\
pongpisit.tha@gmail.com}

\and

\IEEEauthorblockN{Ken-ichi Fukui, Masayuki Numao}
\IEEEauthorblockA{Institute of Scientific and\\
Industrial Research\\
Osaka University\\
\{fukui, numao\}@ai.sanken.osaka-u.ac.jp}

\and

\IEEEauthorblockN{Boonserm Kijsirikul}
\IEEEauthorblockA{Department of Computer Engineering\\
Faculty of Engineering\\
Chulalongkorn University\\
boonserm.k@chula.ac.th}}

\maketitle

%%%%%%%%% ABSTRACT
\begin{abstract}
In this paper, we propose two modified neural networks based on dual path multi-scale fusion networks (SFANet) and SegNet for accurate and efficient crowd counting. Inspired by SFANet, the first model, which is named M-SFANet, is attached with atrous spatial pyramid pooling (ASPP) and context-aware module (CAN). The encoder of M-SFANet is enhanced with ASPP containing parallel atrous convolutional layers with different sampling rates and hence able to extract multi-scale features of the target object and incorporate larger context. To further deal with scale variation throughout an input image, we leverage the CAN module which adaptively encodes the scales of the contextual information. The combination yields an effective model for counting in both dense and sparse crowd scenes. Based on the SFANet decoder structure, M-SFANet's decoder has dual paths, for density map and attention map generation. The second model is called M-SegNet, which is produced by replacing the bilinear upsampling in SFANet with max unpooling that is used in SegNet. This change provides a faster model while providing competitive counting performance. Designed for high-speed surveillance applications, M-SegNet has no additional multi-scale-aware module in order to not increase the complexity. Both models are encoder-decoder based architectures and are end-to-end trainable. We conduct extensive experiments on five crowd counting datasets and one vehicle counting dataset to show that these modifications yield algorithms that could improve state-of-the-art crowd counting methods. Codes are available at \url{https://github.com/Pongpisit-Thanasutives/Variations-of-SFANet-for-Crowd-Counting}.
\end{abstract}

%%%%%%%%% BODY TEXT
\section{Introduction}

Crowd counting is an important task due to its wide range of applications such as public safety, surveillance monitoring, traffic control and intelligent transportation. However, it is a challenging computer vision task and not so trivial to efficiently solve the problem at first glance due to heavy occlusion, perspective distortion, scale variation and diverse crowd distribution throughout real-world images. These problems are emphasized especially when the target objects are in a crowded space. Some of the early methods \cite{dollar2011pedestrian} treat crowd counting as a detection problem. The handcrafted features from multiple sources are also investigated in \cite{idrees2013multi}. These approaches are not suitable when the targeted objects are overlapping each other and the handcrafted features can not handle the diversity of crowd distribution in input images properly. In order to take characteristics of the crowd distribution into account, one should not consider developing models predicting only the number of people in the target image because the characteristics are neglected. Thus, more recent methods rely more on the density map automatically generated from the head annotation ground truth instead. On the other hand, the authors of \cite{ma2019bayesian} consider the density map as a likelihood describing ``how the spatial pixels would be'' given the annotation ground truth and propose the novel Bayesian loss. However, in our experiments, we consider the density map ground truth as the learning target (except our experiment on UCF-QNRF \cite{idrees2018composition}) in order to investigate the improved accuracy caused by the proposed architecture modifications comparing with most of the state-of-the-art methods.

Convolutional Neural Networks (CNNs) have been utilized to estimate accurate density maps. By considering convolutional filters as sliding windows, CNNs are capable of feature extraction throughout various regions of an input image. Consequently, the diversity of crowd distribution in the image is handled more properly. In order to cope with head scale variation problems caused by camera diverse perspectives, previous works mostly make use of multi-column/multi-resolution based architectures \cite{zhang2016single, onoro2016towards, cao2018scale, wu2019adaptive}. However, the study in \cite{li2018csrnet} shows that the features learned at each column structure of MCNN \cite{zhang2016single} are nearly identical and it is not efficient to train such architecture when networks go deeper. As opposed to the multi-column network architecture, a deep single column network based on a truncated VGG16 \cite{simonyan2014very} feature extractor and a decoder with dilated convolutional layers is proposed in \cite{li2018csrnet} and carries out the breakthrough counting performance on ShanghaiTech \cite{zhang2016single} dataset. The proposed architecture demonstrates the strength of VGG16 encoder, which is pretrained on ImageNet \cite{deng2009imagenet} for higher semantics information extraction and the ability to transfer knowledge across vision tasks. Moreover, the study demonstrates how to attach atrous convolutional layers to the network instead of adding more pooling layers which could cause loss of spatial information. Nevertheless, \cite{liu2019context} has raised the issue of using the same filters and pooling operations over the whole image. The authors of \cite{liu2019context} have pointed out that the receptive field size should be changed across the image due to perspective distortion. To deal with this problem, the scale-aware contextual module, named CAN, capable of feature extraction over multiple receptive field sizes, has been proposed in \cite{liu2019context}. By the module design, the importance of each such extracted feature at every image location is learnable. However, CAN contains no mechanism to reduce the background noise, which possibly causes false predictions, especially when facing sparse and complex scenes.

Aside from crowd counting, objects overlapping is also a crucial problem for image segmentation. As a result, scale-aware modules such as spatial pyramid pooling (SPP) \cite{he2015spatial} and atrous spatial pyramid pooling (ASPP) \cite{chen2017deeplab} are devised to capture the contextual information at multiple scales. By employing values of the atrous rate from small to large, the model's field-of-view is enlarged and hence enables object encoding at multiple scales \cite{chen2017deeplab}. Encoder-decoder based CNNs \cite{ronneberger2015u, badrinarayanan2017segnet, chen2018encoder} have prior success in image segmentation ascribed to the ability to reconstruct precise object boundaries. Encoder-decoder based networks have also been proposed for crowd counting such as \cite{cao2018scale, wu2019adaptive, li2018csrnet, wang2019learning, liu2019adcrowdnet, xiong2019open}. Bridging the gap between image segmentation and crowd counting, \cite{zhu2019dual} introduces dual path multi-scale fusion networks (SFANet) which integrate UNet\cite{ronneberger2015u}-like decoder with the dual path structure to predict the head regions among complicated background then regress for the headcounts. Unfortunately, unlike CAN \cite{liu2019context}, SFANet has no explicit module to cope with the scale variation problem.

To redress the weaknesses of the state-of-the-art methods, we propose two modified networks for crowd counting. The first proposed model is called ``M-SFANet'' (Modified-SFANet) in which the multi-scale-aware modules, CAN and ASPP, are additionally connected to the VGG16-bn \cite{simonyan2014very} encoder of SFANet at different encoding stages to resolve the mentioned SFANet’s problem and handle occlusion by capturing the contexts around the targeted objects at multiple scales. As a result of SFANet's dual path decoder for suppressing noisy background, the CAN’s problem is seamlessly cured as well. M-SFANet learns the contextual scales adaptively (following the perspective) and statically; therefore, the model becomes effective for both dense and sparse scenes. Second, we integrate the dual path structure into ModSegNet\cite{ganaye2018semi}-like encoder-decoder network instead of Unet and call this model ``M-SegNet'' (Modified-SegNet). Designed for medical image segmentation, ModSegNet is similar to UNet but leverage max-unpooling \cite{badrinarayanan2017segnet} instead of transpose convolution which is not parameter-free. To the best of our knowledge, the max unpooling is not yet employed in the literature about crowd counting. M-SegNet is designed to be computationally faster (See Table \ref{table:inferspeed}.) than SFANet, while providing the similar performance. Furthermore, we also test the performance of the ensemble model between M-SegNet and M-SFANet by average prediction. For some surveillance applications which speed are not the constraint, the ensemble model should be considered because of its lower variance prediction.
\label{intro_para}

%------------------------------------------------------------------------ 
\section{Related Works}
% Solutions to crowd counting are classified into traditional approaches or CNN-based approaches. The traditional approaches include detection-based methods and regression-based methods. Recently, the CNN-based approaches often outperform the traditional approaches.

%------------------------------------------------------------------------- 
\subsection{Traditional Approaches}
Some early proposals rely on sliding window-based detection algorithms. This requires the extracted feature from human heads or human bodies such as histogram oriented gradients (HOG) \cite{dalal2005histograms}. Unfortunately, these methods fail to detect people when encountering images with high occlusion. Regression-based methods attempt to learn the mapping function from low-level information \cite{chan2009bayesian} generated by features such as foreground and texture to the number of targeted objects. Mapping functions have also been studied in \cite{lempitsky2010learning}.

%------------------------------------------------------------------------- 

\subsection{CNN-based Approaches}
Lately, CNN-based methods have shown significant improvements from traditional methods on the task. Zhang \textit{et al}. \cite{zhang2015cross} proposed a deep trained CNN to estimate crowd count while predicting crowd density level. In \cite{zhang2016single}, Zhang \textit{et al}. proposed a multi-column CNN (MCNN) in which each column was designed to respond to different scales. Li \textit{et al}. \cite{li2018csrnet} proposed a deep single column CNN based on a truncated VGG16 encoder and dilated convolutional layers as a decoder to aggregate the multi-scale contextual information. Cao \textit{et al}. \cite{cao2018scale} presented an encoder-decoder scale aggregation network (SANet). Jiang \textit{et al}. \cite{jiang2019crowd} proposed a trellis encoder-decoder network (TEDnet) that incorporates multiple decoding paths. Shi \textit{et al}. \cite{shi2019revisiting} exploited the perspective information of pedestrian height, which is used to combine the multi-scale density maps. Liu \textit{et al}. \cite{liu2019adcrowdnet} applied an attention mechanism to crowd counting by integrating an attention-aware network into a multi-scale deformable network to detect crowd regions. Wang \textit{et al}. \cite{wang2019learning} escalated the performance of crowd counting by pretraining a crowd counter on the synthetic data and proposed a crowd counting method via domain adaptation to deal with the lack of labeled data. Liu \textit{et al}. \cite{liu2019context} proposed a VGG16-based model with the scale-aware contextual structure (CAN) that combines features extracted from multiple receptive field sizes and learns the importance of each such feature over image location. Zhu \textit{et al}. \cite{zhu2019dual} proposed the dual path multi-scale fusion networks (SFANet) with an additional path to supervisedly learn generated attention maps. SFANet's decoder reuses coarse features and high-level features from encoding stages similar to Unet \cite{ronneberger2015u}. Similarly, Sindagi \textit{et al}. \cite{sindagi2019ha} also introduced the attention mechanism, but embedded as the spatial attention module (SAM) in the network design. Sam \textit{et al}. \cite{sam2018top} delivered a separate top-down feedback CNN to correct the initial prediction. Likewise in \cite{sindagi2019multi}, Sindagi \textit{et al}. designed bottom-top and top-bottom feature fusion method to utilize information present at each of the network layers. Recently, instead of employing density maps as learning targets, Ma \textit{et al}. \cite{ma2019bayesian} constructed a density contribution model and trained a VGG19\cite{simonyan2014very}-based network using Bayesian loss rather than the vanilla mean square error (MSE) loss.
\label{related_work_para}
\section{Proposed Approach}
Since there are two base neural network architectures we modify and experiment with, SFANet \cite{zhu2019dual} and SegNet \cite{badrinarayanan2017segnet}, we coin our models ``M-SFANet'' (Modified SFANet) and ``M-SegNet'' (Modified SegNet) respectively. Both of them are encoder-decoder based deep convolutional neural networks. They commonly have the convolutional layers of VGG16-bn as the encoder which gradually reduces feature maps size and captures high-level semantics information. In the case of M-SFANet, the features are passed through CAN \cite{liu2019context} module and ASPP \cite{chen2017deeplab} to extract the scale-aware contextual features. Finally, the decoders recover the spatial information to generate the final high-resolution density map. As a result of the combination, M-SFANet is more heavyweight and usually predicts more accurate crowd counts compared to the proposed M-SegNet. Based on SegNet, M-SegNet is lightweight and has no additional multi-scale-aware module. However, M-SegNet can achieve competitive results on some crowd counting benchmarks.

\begin{figure}
    \centering
    \includegraphics[width=82mm]{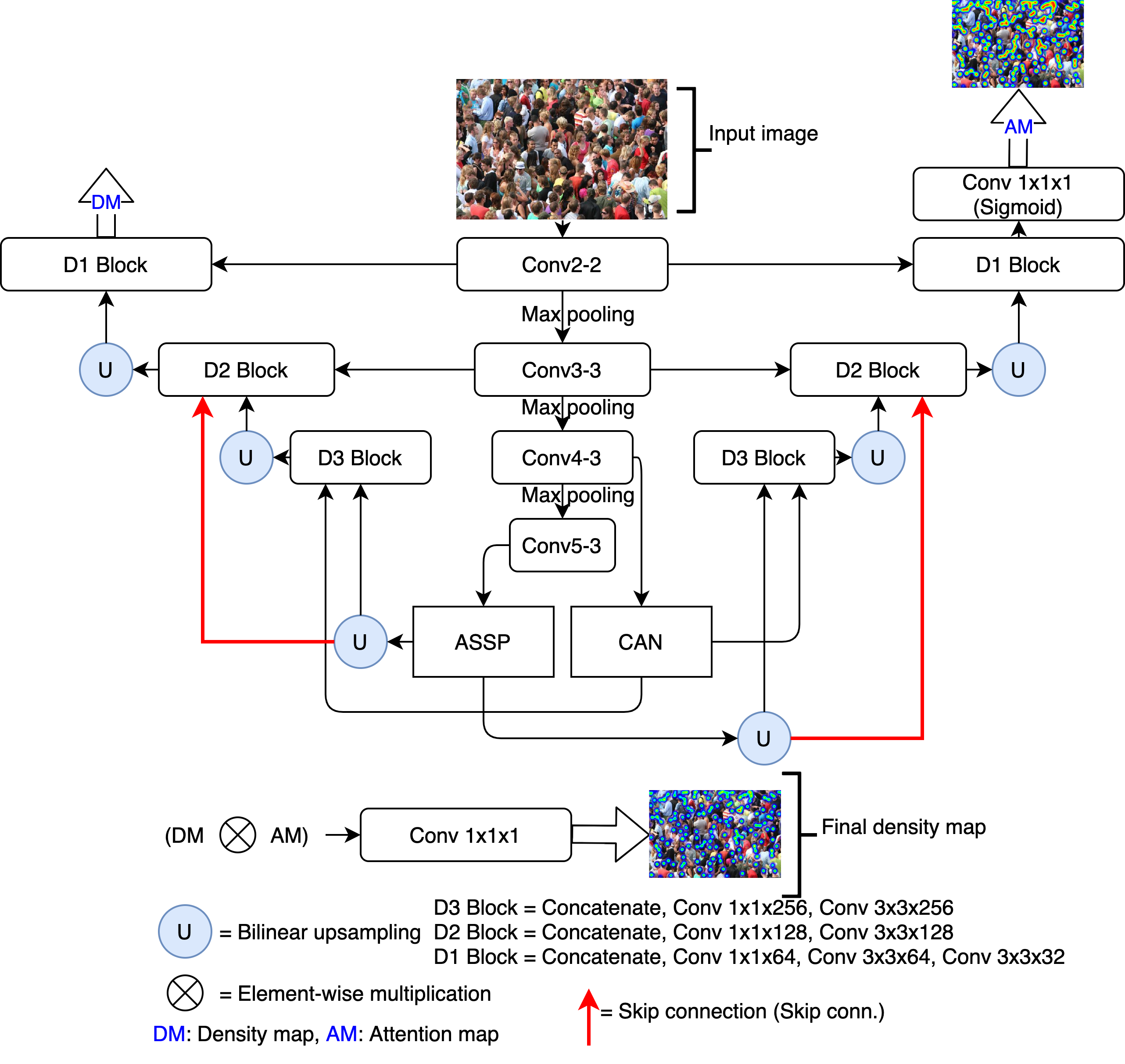}
    \caption{The architecture of M-SFANet. The convolutional layers’ parameters are denoted as Conv (kernel size)-(kernel size)-(number of filters). Max pooling is conducted over a $2\times2$ pixel window with stride 2.}
    \label{fig:M-SFANet architecture}
\end{figure}

\subsection{Modified SFANet (M-SFANet)}
The model architecture consists of 3 novel components, VGG16-bn feature map encoder, the multi-scale-aware modules, and the dual path multi-scale fusing decoder \cite{zhu2019dual}. First, the input images are fed into the encoder to learn useful high-level semantics meaning. Then, the feature maps are fed into the multi-scale-aware modules in order to highlight the multi-scale features of the target objects and the context. There are two multi-scale-aware modules in M-SFANet architecture, one connected with the \nth{13} layer of VGG16-bn is ASPP, and the other module connected with the \nth{10} layer of VGG16-bn is CAN. Lastly, the decoder paths use concatenate and bilinear upsampling to fuse the multi-scale features into the density maps and attention maps. Before producing the final density maps, the crowd regions are segmented from background by the generated attention maps. This mechanism spaces out noise background and lets the model focus more on the regions of interest. We leverage the multi-task loss function in \cite{zhu2019dual} to gain the advantage of the attention map branch. The overview of M-SFANet is shown in Fig. \ref{fig:M-SFANet architecture}. Every convolutional layer is followed by batch normalization \cite{ioffe2015batch} and ReLU except the last convolutional layer for predicting the final density map.

\textbf{Feature map encoder (13 layers of VGG16-bn):} We leverage the first pretrained 13 layers of VGG16 with batch normalization because the stack of 3x3 Convolutional layers is able to extract multi-scale features and multi-level semantic information. This is a more efficient way to deal with scale variation problems than multi-column architecture with different kernel sizes \cite{li2018csrnet}. The high-level feature maps (1/8 in size of the original input) from the \nth{10} layer are fed into CAN to adaptively encode the scales of the contextual information \cite{liu2019context}. Moreover, the top feature maps (1/16 in size of the original input) from the \nth{13} layer are fed into ASPP to statically learn (not specialized for each image location) image-level features (e.g. human head in our experiments) and the contextual information at multiple rates. Only encoding multi-scale information at the top layer is not efficient due to the most shrunken features, which are not appropriate for generating high-quality density maps \cite{li2018csrnet}.

\textbf{Context-aware module:} CAN \cite{liu2019context} module is capable of producing scale-aware contextual features using multiple receptive fields of average pooling operation. Following \cite{liu2019context}, The pooling output scales are 1, 2, 3, 6. The module extracts those features and learns the importance of each such feature at every image location, thus accounting for potentially rapid scale changes within an image \cite{liu2019context}. The importance of the extracted features varies according to the difference from their neighbors. Due to discriminative information fused from different scales, CAN module performs very well when encountering perspective distortion in the crowded scene.

\textbf{Atrous spatial pyramid pooling:} ASPP \cite{chen2017deeplab} module applies several effective fields-of-view of atrous convolution and image pooling to the incoming features, thus capturing multi-scale information. The atrous rates are 1, 6, 12, 18. Thanks to atrous convolution, loss of information related to object boundaries (between human heads and background) throughout convolutional layers in the encoder is alleviated. The field of view of filters is enlarged to incorporate larger context without losing image resolution. Unlike CAN, ASPP treats the importance of the extracted scale features equally across spatial locations; hence, the module is proper for sparse scenes where the contexts are less informative.
% ASPP is empirically proved to be effective for image segmentation in \cite{chen2018encoder}.

\textbf{Dual path multi-scale fusion decoder:} The decoder architecture consists of the density map path and the attention map path as described in \cite{zhu2019dual}. The following strategy is applied to both the density map path and the attention map path. First, the output feature maps from ASPP are upsampled by a factor of 2 using bilinear interpolation and then concatenated with the output feature maps from CAN. Next, the concatenated feature maps are passed through 1x1x256 and 3x3x256 convolutional layers. Again, The fused features are upsampled by a factor of 2 and concatenated with the conv3-3 and the upsampled (by a factor of 4) feature maps from ASPP (depicted as the red connector in Fig. \ref{fig:M-SFANet architecture}) before passing through 1x1x128 and 3x3x128 convolutional layers. This skip connection helps the network remind itself of the learned multi-scale features from high-level image representation. Finally, the 128 fused features are upsampled by a factor of 2 and concatenated with the conv2-2 before passing through 1x1x64, 3x3x64 and 3x3x32 convolutional layers respectively. Because of the use of three upsampling layers, the model can retrieve high-resolution feature maps with 1/2 size of the original input. Element-wise multiplication is applied to the attention map and the density feature maps to generate the refined final density map.
\label{M_SFANet_info}

\begin{figure}
    \centering
    \includegraphics[width=82mm]{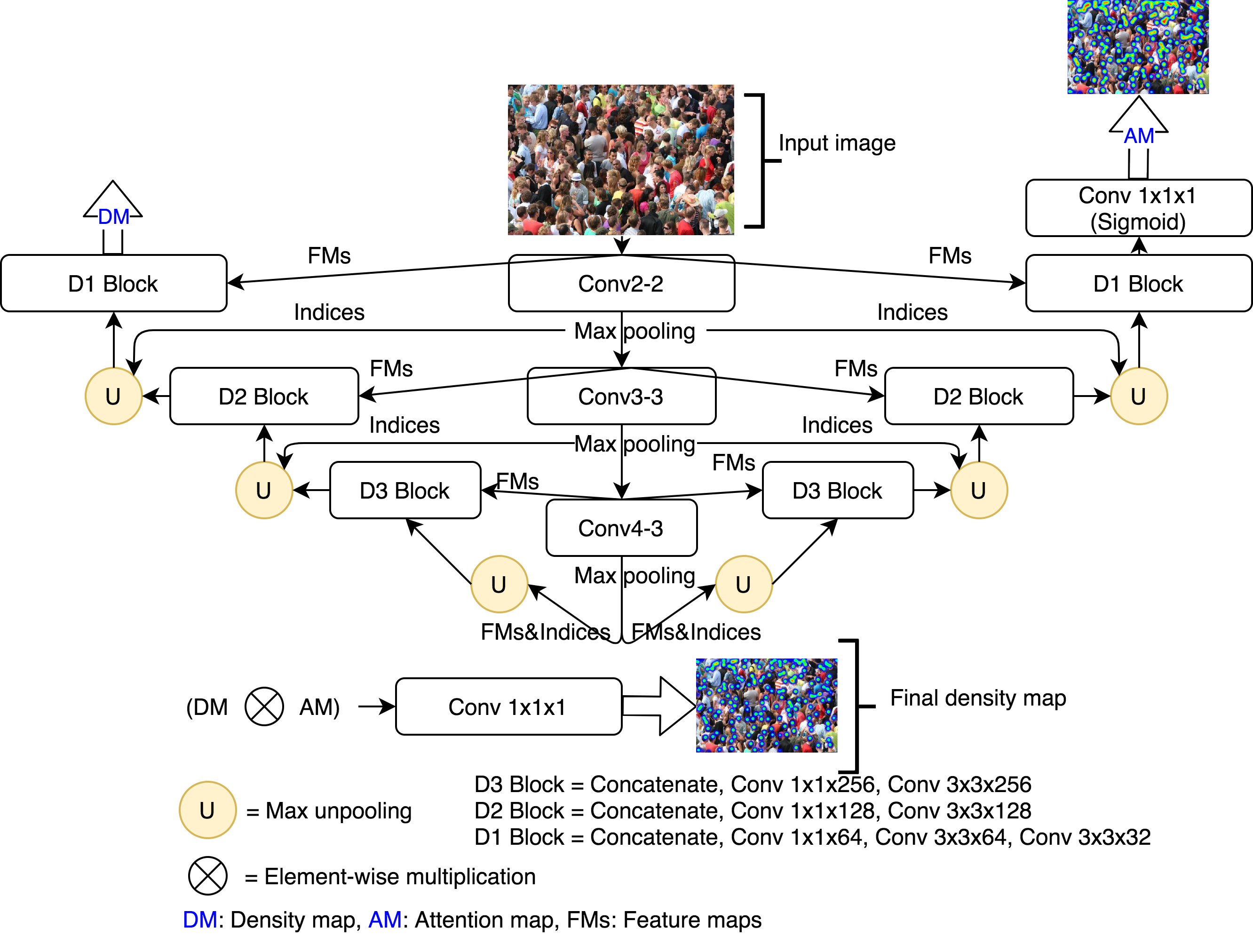}
    \caption{The architecture of the proposed M-SegNet. The convolutional layers' parameters are denoted as Conv (kernel size)-(kernel size)-(number of filters). Max pooling is conducted over a $2\times2$ pixel window with stride 2.}
    \label{fig:M-SegNet architecture}
\end{figure}

\subsection{Modified SegNet (M-SegNet)}
M-SegNet shares almost the same components as presented in M-SFANet except for the fact that there are no CAN module and ASPP to additionally emphasize multi-scale information and the bilinear upsampling is replaced with max unpooling operation using the memorized max-pooling indices \cite{badrinarayanan2017segnet} from the corresponding encoder layer. The first 10 layers of VGG16-bn are employed as the feature map encoder. Hence, M-SegNet requires less computational resources than M-SFANet (See Table \ref{table:params} and Table \ref{table:inferspeed}.) and more suitable for speed-constrained applications. The overview of M-SegNet is shown in Fig. \ref{fig:M-SegNet architecture}.

\section{Training Method}
In this section, we explain how the density map ground truth and the attention map ground truth are automatically generated in our experiments. Training settings for each dataset are shown in Table \ref{table:lr_bs_table}.
\label{training_method_section}

\subsection{Density map ground truth}
To generate the density map ground truth $D(x)$, we follow the Gaussian method with fixed standard deviation kernel described in \cite{zhang2016single}. Assuming that there is a head annotation at pixel $x_{i}$ represented as $\delta(x - x_{i})$, the density map can be constructed by convolution with Gaussian kernel \cite{lempitsky2010learning}. This processes are formulated as:
\begin{equation}
D(x) = \sum_{i=1}^{C} {\delta \left(x - x_{i}\right) * G_{\sigma} (x)}
\end{equation} 
In the ground truth annotation, we convolve each $\delta(x - x_{i})$ with a Gaussian kernel (blurring each head annotation) with parameter $\sigma$, where $C$ is a number of total headcounts. In our experiment we set $\sigma = 5, 4, 4, 10$ for ShanghaiTech \cite{zhang2016single}, UCF\_CC\_50 \cite{bansal2015people}, WorldExpo'10 \cite{zhang2015cross}, and TRANCOS \cite{guerrero2015extremely} dataset. For Beijing BRT \cite{ding2018deeply} dataset, we use the code provided in \url{https://github.com/XMU-smartdsp/Beijing-BRT-dataset} for density map ground truth generation. For UCF\_CC\_50 \cite{bansal2015people} and WE \cite{zhang2015cross}, we borrow the code from \url{https://github.com/gjy3035/C-3-Framework} \cite{gao2019c}.

\subsection{Attention map ground truth}
Following \cite{zhu2019dual}, Attention map ground truth is generated based on the threshold applied to the corresponding density map ground truth. The formulated equation is as follows: 

\begin{equation}
A(\textbf{i}) =
\begin{cases}
      0 & 0.001>D(\textbf{i}) \\
      1 & 0.001\leq D(\textbf{i}) 
   \end{cases}
\end{equation}
where \textbf{i} is a coordinate in the density map ground truth. The threshold is set to 0.001 according to \cite{zhu2019dual}.

\subsection{Training details}
We leverage the image augmentation strategy as described in \cite{zhu2019dual} but the size of the cropped image differs across datasets. Therefore, when training each dataset, we use different learning rates and batch sizes. The main strategy can be summarized as random resizing by small portion, image cropping, horizontal flip, and gamma adjustment. The main difference to \cite{zhu2019dual} is that we use Adam \cite{kingma2014adam} with lookahead optimizer \cite{zhang2019lookahead} to train our models since it shows faster convergence than standard Adam optimizer and experimentally proved to improve the model's performance in \cite{zhang2019lookahead}.

\section{Experimental Evaluation}
In this section, we show the results of our M-SFANet and M-SegNet on 5 challenging crowd counting datasets and 1 congested vehicle counting dataset, TRANCOS \cite{guerrero2015extremely}. We mostly evaluate the performance using mean absolute error (MAE) and root mean square error (RMSE). The metrics are defined as follows:
\begin{align}
  MAE & = \frac{1}{N}\sum_{i=1}^{N} |C^{Prediction}_{i} - C^{Ground}_{i}| \\
  RMSE &= \sqrt{\frac{1}{N}\sum_{i=1}^{N} (C^{Prediction}_{i} - C^{Ground}_{i})^{2}}
\end{align}
where $N$ is the number of test images. $C^{Prediction}_{i}$ and $C^{Ground}_{i}$ refer to the prediction headcounts and the ground truth headcounts for the $i^{th}$ test image.

\setlength{\tabcolsep}{2pt}
\begin{table}
\begin{center}
\caption{Training settings for each dataset}
\begin{tabular}{c|c|c|c}
\hline
\textbf{Dataset}
 & learning rate & batch size & crop size\\
\hline
ShanghaiTech \cite{zhang2016single} & 5e-4 & 8 | 8 & 400x400\\
UCF\_CC\_50 \cite{bansal2015people} & 8e-4 & 5 | 8 & 512x512\\
WE \cite{zhang2015cross} & 8e-4 | 5e-4 & 42 | 45 & 224x224\\
BRT \cite{ding2018deeply} & 6e-4 & 42 | 45 & 224x224\\
UCF-QRNF \cite{idrees2018composition} & 5e-4 & 5 | 5 & 512x512\\
TRANCOS \cite{guerrero2015extremely} & 5e-4 & 5 | 8 & full image**\\
\hline
\end{tabular}
\label{table:lr_bs_table}
\end{center}
\footnotesize\emph{Note:} ``|'' separates batch size for M-SFANet (left) and M-SegNet (right).\\
% *We recommend using as large as possible. We select these values based on the GPU memory limit.\\
**Full image training means no random resize and no image cropping.
\end{table}

\begin{figure}
    \centering
    \includegraphics[width=82mm]{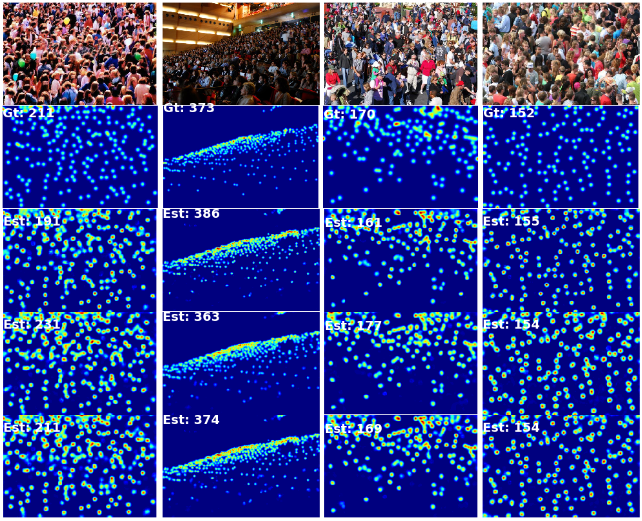}
    \caption{Visualization of estimated density maps. The first row is sample images from ShanghaiTech Part A. The second row is the ground truth. The $3^{rd}$ to $5^{th}$ rows correspond to the estimated density maps from M-SegNet, M-SFANet and M-SegNet+M-SFANet respectively.}
    \label{fig:heatmap_visual_sha}
\end{figure}

\subsection{ShanghaiTech dataset}
ShanghaiTech \cite{zhang2016single} dataset consists of 1198 labeled images with 330,165 annotated people. The dataset is divided into Part A (SHA) and Part B (SHB). SHA contains 482 (train:300, test:182) highly congested images downloaded from the internet. SHB includes 716 (train:400, test:316) relatively sparse crowd scenes taken from streets in Shanghai. Table \ref{table:shbres} shows the results of our models compared to state-of-the-art methods and the ablation study. Compared to the base model, SFANet, M-SFANet can reduce MAE by 2.03\% on SHB. Indicating MAE/RMSE improvement by 1.45\%/4.50\%, M-SegNet is able to outperform SFANet on SHB as well. By average prediction between M-SFANet and M-SegNet, we can gain 3.76\% MAE and 8.41\% MAE relative improvement on SHA and SHB respectively. Moreover, M-SFANet and M-SegNet both show competitive results compared to the best methods on SHA (S-DCNet \cite{xiong2019open}) and SHB (SANet+SPANet \cite{cheng2019learning}). According to the ablation study in Table \ref{table:shbres}, M-SFANet w/o CAN attain a higher MAE than M-SFANet w/o ASPP on SHA while the result is converse on SHB. This empirically shows that CAN and ASPP are effective for crowded scenes and sparse scenes respectively. Subsequently, integrating both modules successfully decreases MAE/RMSE on both SHA and SHB. Retaining the useful features extracted from ASPP, the skip connection also enhances the counting performance. In the ablation study, we have found that fine-tuning our M-SFANet* pretrained on UCF-QNRF (described in section \ref{ucfqnrf_section}) using Bayesian loss \cite{ma2019bayesian} is a viable transfer learning approach to achieve accurate counting on SHB. The visualization of estimated density maps by our models on SHA is shown in Fig. \ref{fig:heatmap_visual_sha}.
\label{SH_discussion}

\setlength{\tabcolsep}{4pt}
\begin{table}
\begin{center}
\caption{Comparison with state-of-the-art methods on ShanghaiTech \cite{zhang2016single} dataset and ablation study of M-SFANet}
\begin{tabular}{c|c|c|c|c}
\hline
Method & \multicolumn{2}{c|}{Part A} & \multicolumn{2}{c}{Part B}\\ \cline{2-5}
 & MAE & RMSE & MAE & RMSE\\
\hline
MCNN \cite{zhang2016single} & 110.2 & 173.2 & 26.4 & 41.3\\
CSRNet \cite{li2018csrnet} & 68.2 & 115.0 & 10.6 & 16.0\\
TEDNet \cite{jiang2019crowd} & 64.2 & 109.1 & 8.2 & 12.8\\
CAN \cite{liu2019context} & 62.3 & 100.0 & 7.8 & 12.2\\
HA-CCN \cite{sindagi2019ha} & 62.9 & 94.9 & 8.1 & 13.4\\
SFANet \cite{zhu2019dual} & 59.8 & 99.3 & 6.9 & 10.9\\
BL \cite{ma2019bayesian} & 62.8 & 101.8 & 7.7 & 12.7\\
MBTTBF-SCFB \cite{sindagi2019multi} & 60.2 & 94.1 & 8.0 & 15.5\\
S-DCNet \cite{xiong2019open} & 58.3 & 95.0 & 6.7 & 10.7\\
SANet+SPANet \cite{he2015spatial} & 59.4 & \textbf{92.5} & 6.5 & \textbf{9.9}\\
\hline
M-SFANet w/o CAN \cite{liu2019context} & 62.41 & 101.13 & 7.40 & 12.14\\
M-SFANet w/o ASPP \cite{chen2017deeplab} & 61.25 & 102.37 & 7.67 & 13.28\\
M-SFANet w/o skip conn. & 60.07 & 99.47 & 7.34 & 12.10\\
M-SFANet* & 62.49 & 106.11 & 6.38 & 10.22\\
\hline
M-SegNet & 60.55 & 100.80 & 6.80 & 10.41\\
M-SFANet & 59.69 & 95.66 & 6.76 & 11.89\\
M-SFANet+M-SegNet & \textbf{57.55} & 94.48 & \textbf{6.32} & 10.06\\
\hline
\end{tabular}
\label{table:shbres}
\end{center}
\footnotesize\emph{Note:} M-SFANet* denotes fine-tuning the pretrained model from section \ref{ucfqnrf_section} using the Bayesian loss \cite{ma2019bayesian}. ``M-SFANet+M-SegNet'' determines average prediction between the two models. The best performance is on boldface.
\end{table}
\setlength{\tabcolsep}{1.4pt}
\setlength{\tabcolsep}{4pt}
\begin{table}
\begin{center}
\caption{Comparison with state-of-the-art methods on UCF\_CC\_50 \cite{bansal2015people}}
\begin{tabular}{c|c|c}
\hline
Method & MAE & RMSE\\
\hline
CSRNet \cite{li2018csrnet} & 266.1 & 397.5\\
TEDNet \cite{jiang2019crowd} & 249.4 & 354.5\\
CAN \cite{liu2019context} & 212.2 & \textbf{243.7}\\
HA-CCN \cite{sindagi2019ha} & 256.2 & 348.4\\
SFANet \cite{zhu2019dual} & 219.6 & 316.2\\
BL \cite{ma2019bayesian} & 229.3 & 308.2\\
MBTTBF-SCFB \cite{sindagi2019multi} & 233.1 & 300.9\\
SANet+SPANet \cite{he2015spatial} & 232.6 & 311.7\\
S-DCNet \cite{xiong2019open} & 204.2 & 301.3\\
\hline
M-SegNet & 188.40 & 262.21\\
M-SFANet & \textbf{162.33} & 276.76\\
M-SFANet+M-SegNet & 167.51 & 256.26\\
\hline
\end{tabular}
\label{table:ucfres}
\end{center}
\footnotesize\emph{Note:} ``M-SFANet+M-SegNet'' determines average prediction between the two models. The best performance is on boldface.
\end{table}
\setlength{\tabcolsep}{1.4pt}

\subsection{UCF\_CC\_50 dataset}
\begin{figure}
    \centering
    \includegraphics[width=82.0mm]{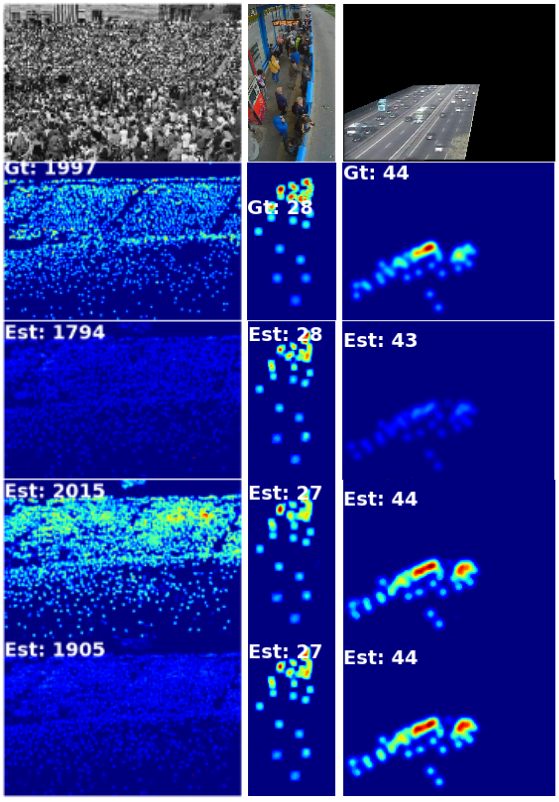}
    \caption{Visualization of estimated density maps. The first row are sample images from UCF\_CC\_50 \cite{bansal2015people}, Beijing BRT \cite{ding2018deeply} and TRANCOS \cite{guerrero2015extremely} dataset (from left to right). The second row is the ground truth. The $3^{rd}$ to $5^{th}$ rows correspond to the estimated density maps from M-SegNet, M-SFANet and M-SegNet+M-SFANet respectively.}
    \label{fig:ucfcc_brt_trancos}
\end{figure}

Proposed by \cite{bansal2015people}, the dataset contains extremely crowded scenes with limited training samples. It includes only 50 high-resolution images with numbers of head annotations ranging from 94 to 4543. Because of the limited numbers of training samples, we pretrain our models on ShanghaiTech Part A. To evaluate model performance, 5-fold cross-validation is performed following the standard setting in \cite{bansal2015people}. The results compared to the state-of-the-art methods are listed in Table \ref{table:ucfres}. M-SFANet obtains the competitive performance with 20.5\% MAE improvement compared with the second-best approach, S-DCNet \cite{xiong2019open}. The results indicate that the ensemble model does not always produce superior performance since the two proposed networks are similar in encoder and decoder structure, the trained models could be minor in diversity. The visualization of the predicted density maps on a dense scene of this dataset is depicted in the left column of Fig. \ref{fig:ucfcc_brt_trancos}.
\label{UCF_CC_50_discussion}

\subsection{WorldExpo'10 dataset}
It includes 1,132 annotated video sequences collected from 103 different scenes. There are 3,980 annotated frames and 3,380 of them are used for model training. Each scene has a Region Of Interest (ROI). Having no access to the original dataset, we use the images and the density map ground truth generated by \cite{gao2019c} to train our models. In Table \ref{table:weres}, the performance on each test scene is reported in MAE. M-SegNet and M-SFANet achieve the best performance in scene 1 (sparse crowd) and scene 4 (dense crowd) respectively. 
% Fig. \ref{fig:we_vis} depicts the visualization of predicted density maps from M-SFANet (top) and M-SegNet (bottom).

% \begin{figure}
%     \centering
%     \includegraphics[width=82mm]{etcs/pic/we_vis.png}
%     \caption{Visualization of estimated density maps from M-SFANet ($1^{st}$ row) and M-SegNet ($2^{th}$ row) on test samples of WorldExpo'10 \cite{zhang2015cross} dataset}.
%     \label{fig:we_vis}
% \end{figure}

\setlength{\tabcolsep}{4pt}
\begin{table}
\begin{center}
\caption{Comparison with state-of-the-art methods on WE \cite{zhang2015cross} dataset}
\begin{tabular}{c|c|c|c|c|c|c}
\hline
Method & Sce.1 & Sce.2 & Sce.3 & Sce.4 & Sce.5 & Ave.\\
\hline
MCNN \cite{zhang2016single} & 3.4 & 20.6 & 12.9 & 13.0 & 8.1 & 11.6\\
CSRNet \cite{li2018csrnet} & 2.9 & 11.5 & 8.6 & 16.6 & 3.4 & 8.6\\
CAN \cite{liu2019context} & 2.9 & 12.0 & 10.0 & 7.9 & 4.3 & 7.4\\
PGCNet \cite{yan2019perspective} & 2.5 & 12.7 & \textbf{8.4} & 13.7 & 3.2 & 8.1\\
DSSINet \cite{liu2019crowd} & 1.57 & \textbf{9.51} & 9.46 & 10.35 & \textbf{2.49} & \textbf{6.67}\\
\hline
M-SegNet & \textbf{1.45} & 11.72 & 10.29 & 21.15 & 5.47 & 10.03\\
M-SFANet & 1.88 & 13.24 & 10.07 & \textbf{7.5} & 3.87 & 7.32\\
\hline
\end{tabular}
\label{table:weres}
\end{center}
\footnotesize\emph{Note:} The result of ``M-SFANet+M-SegNet'' is not included because of no improvement from the state-of-the-art-methods. The best performance is on boldface.
\end{table}
\setlength{\tabcolsep}{1.4pt}

\subsection{Beijing BRT dataset}
Beijing BRT dataset \cite{ding2018deeply} is a crowd counting dataset for intelligent transportation. The number of heads varies from 1 to 64. The images are all 640$\times$360 pixels and taken from the Bus Rapid Transit (BRT) in Beijing. The images are taken from morning until night, therefore they contain shadows, glare, and sunshine interference. Table \ref{table:brtres} reports our models' performance on this dataset. M-SFANet+M-SegNet obtains the new best performance with 17.27\%/9.50\% MAE/RMSE relative improvement. The visualization of the estimated density maps on a sample of this dataset is shown in the middle column of Fig. \ref{fig:ucfcc_brt_trancos}.

\setlength{\tabcolsep}{4pt}
\begin{table}
\begin{center}
\caption{Comparison with state-of-the-art methods on Beijing BRT \cite{ding2018deeply} dataset}
\begin{tabular}{c|c|c}
\hline
Method & MAE & RMSE\\
\hline
MCNN \cite{zhang2016single} & 2.24 & 3.35\\
% FCN \cite{ding2018deeply} & 1.74 & 2.43\\
ResNet-14 \cite{ding2018deeply} & 1.48 & 2.22\\
DR-ResNet \cite{ding2018deeply} & 1.39 & 2.00\\
\hline
M-SegNet & 1.26 & 1.98\\
M-SFANet & 1.16 & 1.90\\
M-SFANet+M-SegNet & \textbf{1.15} & \textbf{1.81}\\
\hline
\end{tabular}
\label{table:brtres}
\end{center}
\footnotesize\emph{Note:} ``M-SFANet+M-SegNet'' determines average prediction between the two models. The best performance is on boldface.
\end{table}
\setlength{\tabcolsep}{1.4pt}

\subsection{UCF-QNRF dataset}
\begin{figure}
    \centering
    \includegraphics[width=82mm]{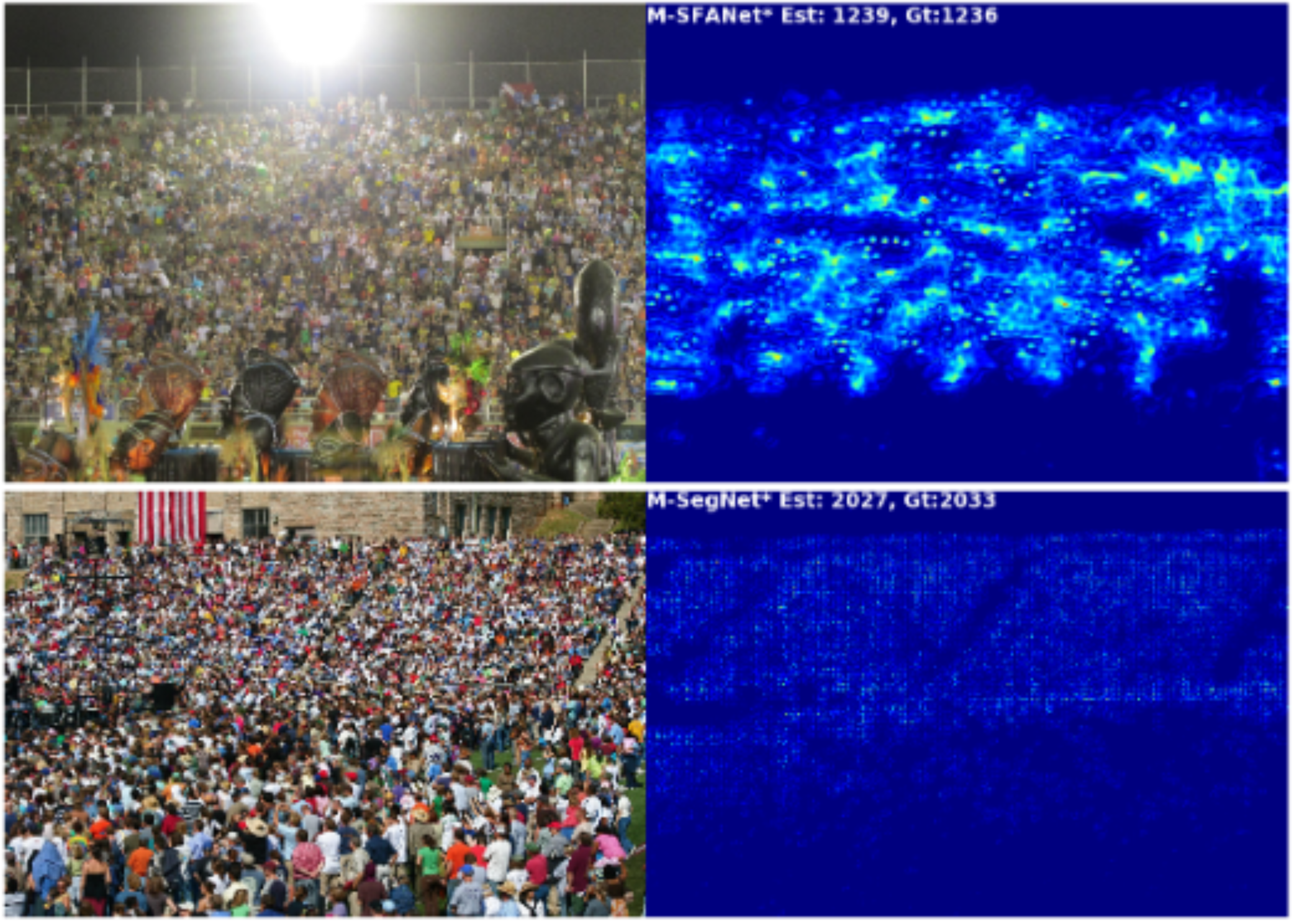}
    \caption{Visualization of estimated density maps from M-SFANet* ($1^{st}$ row) and M-SegNet* ($2^{nd}$ row) on test images of UCF-QNRF \cite{idrees2018composition} dataset.}
    \label{fig:ucfqnrf}
\end{figure}

UCF-QNRF \cite{idrees2018composition} is a new and challenging dataset due to the extremely congested scenes. The dataset contains 1535 high-resolution (2013×2902 on average) images (train:1201, test:334) with 1.25M head annotations. We leverage the training scheme described in BL \cite{ma2019bayesian} except the crowd counter architecture and the optimizer. Following the successful use of VGG19 \cite{simonyan2014very} backbone in BL, we replace the encoder of M-SFANet and M-SegNet with 16 layers of VGG19, input the BL decoder's output feature maps (128 channels) to the D3 block and remove the attention map branch to reduce the complexity. The results are reported in Table \ref{table:ucfqnrf}. Upgraded from BL, M-SFANet* is able to push the current state-of-the-art performance. Fig \ref{fig:ucfqnrf}. depicts the density map visualization.
\label{ucfqnrf_section}

\setlength{\tabcolsep}{4pt}
\begin{table}
\begin{center}
\caption{Comparison with state-of-the-art methods on UCF-QNRF \cite{idrees2018composition}}
\begin{tabular}{c|c|c}
\hline
Method & MAE & RMSE\\
\hline
TEDNet \cite{jiang2019crowd} & 113.0 & 188.0\\
CAN \cite{liu2019context} & 107.0 & 183.0\\
HA-CCN \cite{sindagi2019ha} & 118.1 & 180.4\\
SFCN \cite{wang2019learning} & 102.0 & 171.4\\
SFANet \cite{zhu2019dual} & 100.8 & 174.5\\
S-DCNet \cite{xiong2019open} & 104.4 & 176.1\\
DSSINet \cite{liu2019crowd} & 99.1 & 159.2\\
BL \cite{ma2019bayesian} & 88.7 & 154.8\\
MBTTBF-SCFB \cite{sindagi2019multi} & 97.5 & 165.2\\
\hline
M-SegNet* & 95.17 & 153.73\\
M-SFANet* & \textbf{85.60} & 151.23\\
M-SFANet*+M-SegNet* & 87.64 & \textbf{147.78}\\
\hline
\end{tabular}
\label{table:ucfqnrf}
\end{center}
\footnotesize\emph{Note:} * refers to the modifications described in section \ref{ucfqnrf_section}. ``M-SFANet*+M-SegNet*'' determines average prediction between the two models. The best performance is on boldface.
\end{table}
\setlength{\tabcolsep}{1.4pt}

\subsection{TRANCOS dataset}
Apart from crowd counting, we also evaluate our models on TRANCOS \cite{guerrero2015extremely}, a vehicle counting dataset, to demonstrate the robustness and generalization of our approaches. The dataset contains 1244 images of different congested traffic scenes taken by surveillance cameras. Each image has a region of interest (ROI) used for evaluation. Following the work in \cite{guerrero2015extremely}, we use Grid Average Mean Absolute Error (GAME) for model performance evaluation. The metric is defined in equation \ref{eq:game}. Our approaches, especially M-SFANet, surpass the best previous method as shown in Table \ref{table:tranres}. The results show that the average density map estimation improves counting accuracy (reduce MAE) but does not provide better localization of target objects. The generated density maps from our models are shown in the right column of Fig. \ref{fig:ucfcc_brt_trancos}.

\begin{align}
  GAME(L) & = \frac{1}{N}\sum_{i=1}^{N}\sum_{l=1}^{4^{L}}|C^{Prediction}_{l,i} - C^{Ground}_{l,i}|\;
\label{eq:game}
\end{align}

\setlength{\tabcolsep}{2pt}
\begin{table}
\begin{center}
\small
\caption{Comparison with state-of-the-art methods on TRANCOS \cite{guerrero2015extremely}}
\begin{tabular}{c|c|c|c|c}
\hline
Method & MAE & GAME(1) & GAME(2) & GAME(3)\\
\hline
CSRNet \cite{li2018csrnet} & 3.56 & 5.49 & 8.57 & 15.04\\
ADCrowdNet \cite{liu2019adcrowdnet} & 2.44 & 4.14 & 6.78 & 13.58\\
S-DCNet \cite{xiong2019open} & 2.92 & 4.29 & 5.54 & 7.05\\
\hline
M-SegNet & 2.51 & 5.43 & 7.59 & 9.49\\
M-SFANet & 2.23 & \textbf{3.46} & \textbf{4.86} & \textbf{6.91}\\
M-SFANet+M-SegNet & \textbf{2.22} & 3.87 & 5.51 & 7.37\\
\hline
\end{tabular}
\label{table:tranres}
\end{center}
\footnotesize\emph{Note:} ``M-SFANet+M-SegNet'' determines average prediction between the two models. MAE is equivalent to GAME(0). The best performance is on boldface.
\end{table}
where N corresponds to the number of test images. $C^{Prediction}_{l,i}$ and $C^{Ground}_{l,i}$ are the predicted and ground truth counts of the $l^{th}$ sub-region of $i^{th}$ test image.

\setlength{\tabcolsep}{4pt}
\begin{table}
\begin{center}
\caption{The parameters and MACs of our models}
\begin{tabular}{c|c|c|c}
\hline
Method & \#Parameters(M) & MACs(G) & Training time (sec/batch)\\
\hline
CAN \cite{liu2019context} & 18.10 & 21.99 & 0.28\\
SFANet \cite{zhu2019dual} & 17.00 & 19.94 & 0.53\\
\hline
M-SegNet & \textbf{\textcolor{red}{9.75}} & \textbf{\textcolor{red}{18.14}} & 0.46\\
M-SFANet & 22.90 & 22.05 & 1.06\\
M-SegNet* & 22.58 & 22.79 & \textbf{\textcolor{red}{0.27}}\\
M-SFANet* & 28.62 & 25.08 & 0.36\\
\hline
\end{tabular}
\label{table:params}
\end{center}
\footnotesize\emph{Note:} * refers to the modifications described in section \ref{ucfqnrf_section}. The parameters and MACs are computed with the input size of 224×224. The training time is the average time to train one batch (8 samples) of ShanghaiTech part B \cite{zhang2016single}.
\end{table}
\setlength{\tabcolsep}{1.4pt}
\setlength{\tabcolsep}{4pt}
\begin{table}
\begin{center}
\caption{Inference time for different resolutions in msec}
\begin{tabular}{|c|c|c|c|c|}
\hline
Method & \multicolumn{4}{c|}{Inference speed (ms)}\\\cline{2-5}
& 320×240 & 640×480 & 1280×960 & 1600×1200\\
\hline
CAN \cite{liu2019context}&15.9&54.9&226.0&370.7\\
SFANet \cite{zhu2019dual}&16.6&60.2&245.3&365.5\\
\hline
M-SegNet&\textbf{\textcolor{red}{13.6}}&\textbf{\textcolor{red}{51.8}}&\textbf{\textcolor{red}{208.7}}&349.4\\
M-SFANet&20.0&70.9&263.6&392.9\\
M-SegNet*&15.1&51.99&211.5&\textbf{\textcolor{red}{332.6}}\\
M-SFANet*&20.0&68.0&270.3&418.6\\
\hline
\end{tabular}
\label{table:inferspeed}
\end{center}
\footnotesize\emph{Note:} * refers to the modifications described in section \ref{ucfqnrf_section}. Each value is the average result from 100 runs on a single Google Cloud's NVIDIA Tesla V4 GPU.
\end{table}
\setlength{\tabcolsep}{1.4pt}

\section{Conclusions}
In this paper, we propose two modified end-to-end trainable neural networks, named M-SFANet and M-SegNet by combining novel architectures, designed for crowd counting and image segmentation. For M-SFANet, we add the multi-scale-aware modules to SFANet architecture for better tackling drastic scale changes of target objects. The model alleviates the drawbacks present in the state-of-the-art methods and therefore shows superior performance on both crowd and vehicle counting. Furthermore, the decoder structure of M-SFANet is adjusted to have more residual connections in order to ensure that the learned multi-scale features of high-level semantic information will impact how the model regress for the final density map. However, the sampling rates of the scale-aware modules are not learnable and the number of these rates is fixed before training. This could lead to limited performance in certain unseen scenes. Hence, an adaptive implementation of the modules in which the sampling rates or dilated rates are adjustable is considered as possible future work. For M-SegNet, we change the up-sampling algorithm from bilinear to max unpooling using the memorized indices employed in SegNet. This yields the cheaper computation model while providing competitive counting performance applicable to real-world applications.
\label{conclusion_para}

{\small
\bibliographystyle{ieeetr}
\bibliography{root}
}

\end{document}